\documentclass[10pt, conference, compsocconf]{IEEEtran}
\usepackage{lineno,hyperref}
\usepackage{booktabs}
\usepackage{multirow}
\usepackage{comment, balance}
\usepackage{amsmath}
\usepackage{subfig} 
\usepackage{csquotes}
\usepackage{xcolor}
\usepackage{balance}
\ifCLASSINFOpdf
   \usepackage[pdftex]{graphicx}
\else
\fi
\usepackage{url}

\begin{document}
%
\title{Few-Shot Learning by Integrating Spatial and Frequency Representation}


\author{\IEEEauthorblockN{Xiangyu Chen$^{\dag}$, Guanghui Wang$^{\ddag}$}
\IEEEauthorblockA{$^\dag$ \textit{Department of Electrical Engineering and Computer Science, University of Kansas, Lawrence KS, USA, 66045}\\
$^\ddag$ \textit{Department of Computer Science, Ryerson University, Toronto ON, Canada, M5B 2K3}\\
}
}


%


\maketitle

\begin{abstract}
Human beings can recognize new objects with only a few labeled examples, however, few-shot learning remains a challenging problem for machine learning systems. Most previous algorithms in few-shot learning only utilize spatial information of the images. In this paper, we propose to integrate the frequency information into the learning model to boost the discrimination ability of the system. We employ Discrete Cosine Transformation (DCT) to generate the frequency representation, then, integrate the features from both the spatial domain and frequency domain for classification. The proposed strategy and its effectiveness are validated with different backbones, datasets, and algorithms. Extensive experiments demonstrate that the frequency information is complementary to the spatial representations in few-shot classification. The classification accuracy is boosted significantly by integrating features from both the spatial and frequency domains in different few-shot learning tasks. 
\end{abstract}


%
\IEEEpeerreviewmaketitle

\section{Introduction}
With the explosion in the amount of data we generate today, deep learning, as a data-driven method, has become the hotspot and achieved significant performance in many directions in computer vision  \cite{he2016deep,cen20deep,wu2021,he2021,ma2020,sajid2020}.  However, there are still many scenarios when we cannot access enough training data, especially in the medical field. For instance, to classify rare diseases, we have only limited data to build models. In addition, collecting a large scale of data could be expensive in some cases. Compared to the data-driven algorithms, human beings can learn a visual concept even using only a single example  \cite{schmidt2009meaning}. This challenge gives birth to a new topic in computer vision, few-shot learning (FSL).

Exploiting the successful application with large-scale data, a direct way to improve few-shot learning performance is through data-augmentation \cite{liu2019few,zhang2019few,chen2019image,wang2018low}.  By feeding the network with more positive and/or negative samples, it may mimic large-scale data tasks and achieve a reasonable performance. Although data augmentation could be an auxiliary method for few-shot learning algorithms, adding more data could not solve the few-shot learning problem essentially since it is impossible to generate \enquote{enough} meaningful training data (like thousands of data) to cover as much as possible different distributions of the data and turn it into a large-scale data problem.

The task of few-shot learning is to recognize novel classes with only a few (e.g. up to 5 \cite{vinyals2016matching}) given labeled images. Many researchers started to focus on using the task-level method \cite{finn2017model,ravi2016optimization,wang2018low,sung2018learning}, meta-learning \cite{thrun2012learning}, to solve the few-shot learning problem. Meta-learning, or \enquote{learn to learn \cite{thrun2012learning}}, treats every classification task as a single task. For example, for $k$-way $n$-shot image classification where each task contains images from $k$ classes with $n$ labeled images for each class, the goal is to recognize the test images with only $k\times n$ labeled images for every task. The $k$-way $n$-shot image set can be regarded as one classification task, which is also called \enquote{episode} \cite{vinyals2016matching} Meta-learning aims to learn transferable knowledge from past experience and the study can be divided as following three categories \cite{finn2017model}, memory-based method \cite{yoo2019coloring,santoro2016meta}, metric-based method \cite{vinyals2016matching,sung2018learning,snell2017prototypical}, and optimization-based method \cite{finn2017model,ravi2016optimization}. Memory-based methods endeavor to use external memories \cite{yoo2019coloring,santoro2016meta} to memorize past information. Optimization-based methods learn to optimize \cite{finn2017model}. The metric-based methods learn to compare between a few labeled images and the test images \cite{vinyals2016matching,sung2018learning,snell2017prototypical}. Nevertheless, all methods mentioned above use only spatial images as the input of the backbone.
\begin{figure*}[tp]
\centering
\includegraphics[width=14cm,height=5 cm]{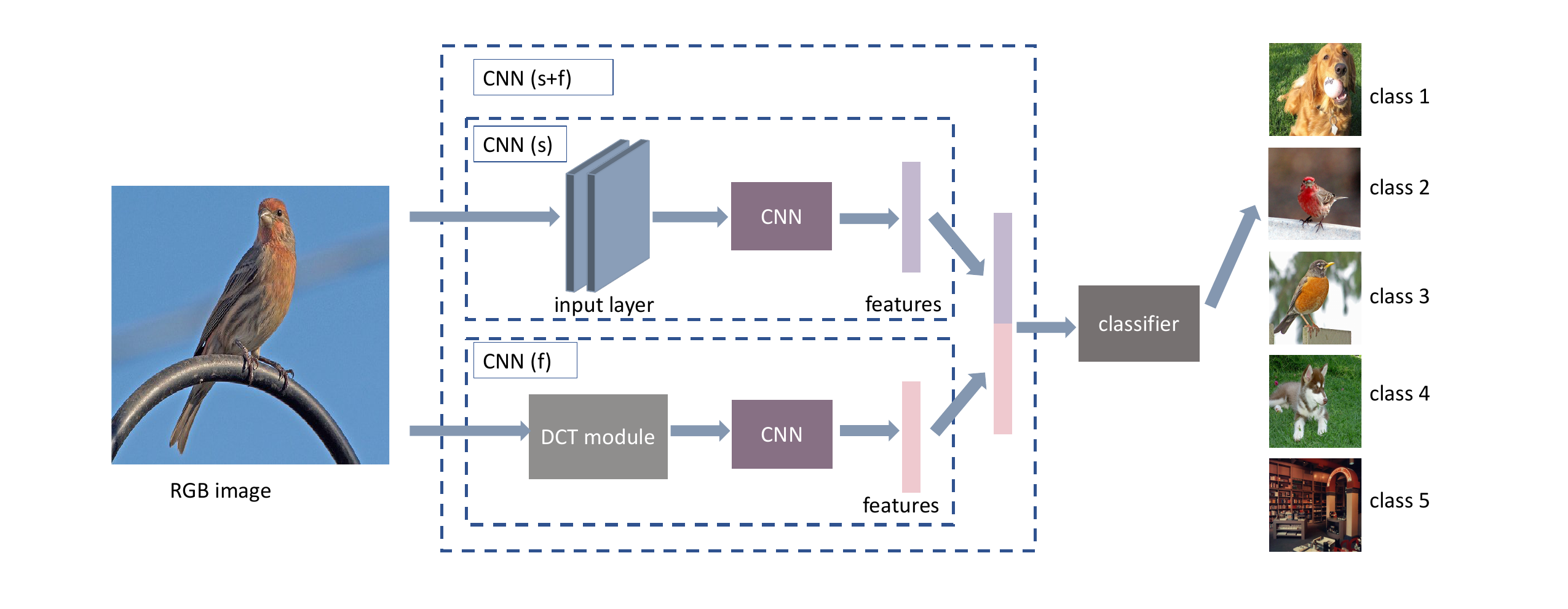}%
\caption{The structure of the proposed framework. We concatenate the features from two networks. The upper network, \enquote{CNN (s)} denotes a regular image classification network where we draw the input layers before the backbone from the whole CNN and \enquote{(s)} means the \enquote{spatial domain}. In the lower network, \enquote{CNN (f)}, the images will go through the DCT module first to generate the frequency representation before being supplied to the CNN backbone, and \enquote{(f)} represents the \enquote{frequency domain}. Finally, we concatenate features obtained from both networks to generate the final classification score, which is the whole output network, \enquote{CNN (s+f)}.}
\label{fig1}
\end{figure*}

Inspired by the fact that human vision is more sensitive to low-frequency information {}{\cite{kim2017deep, XuKai2020}}, we assume that human beings are learning some frequency information when classifying images. Results from the recent research \cite{XuKai2020} also demonstrated impressive performance in the frequency domain where the backbone networks are fed with frequency domain information. To exploit the advantages of both the spatial domain and frequency domain information, we propose to integrate the features extracted from the spatial domain and the frequency domain, aiming to increase the performance of few-shot learning. As illustrated in Figure \ref{fig1}, the spatial domain feature module employs CNN to obtain the representation of original images, and the frequency domain feature module utilizes the Discrete Cosine Transformation (DCT) module \cite{ahmed1974discrete} to generate the frequency representation of original images and feed them into CNN to generate frequency features. Both the frequency features and spatial features are then fused together, followed by a classifier. 

The main contributions of this study are as follows: 
\begin{enumerate}
\item The paper proposes to exploit DCT with static selected channels to few-shot classification and implements the idea on different FSL models. 
\item We investigate the influence of DCT filter size on the few-shot classification and find the relationship between DCT filter size and the classification accuracy.
\item Extensive experiments demonstrate that integrating the features from both the spatial and frequency domains can significantly increase the classification accuracy in few-shot classification. 

\end{enumerate}
The source code of the proposed model can be downloaded from  {\url{https://github.com/xiangyu8/PT-MAP-sf}}.

\section{Related work}

\subsection{Few-shot Learning}

Few-shot learning aims to learn to classify query examples from \enquote{novel} classes given a few labeled support examples from the \enquote{novel} classes and abundant labeled support examples from \enquote{base} classes. Recent deep learning based few-shot learning algorithms could be roughly divided into 4 categories: 1) Data augmentation is a direct data-level method to improve classification accuracy in few-shot learning via mimicking large-scale data algorithms \cite{liu2019few,zhang2019few,chen2019image,wang2018low}, which is usually added to meta-learners as auxiliary.  By generating more positive and/or negative data according to the given labeled data, more information could be fed to the deep neural network. 2) Metric-learning based method, or learn to compare, is one type of meta-learning based approaches in few-shot learning,  which focuses on constructing an appropriate embedding space to yield corresponding features of images and then calculating the similarity between the features of given labeled images and test images.  Related researches include  \cite{li2019large,peng2019few,mangla2020charting,li2019few,vinyals2016matching,snell2017prototypical}.  3) Optimization-based meta-learning methods \cite{ravi2016optimization,finn2017model,nichol2018first}, or learn to optimize, usually train another network to get the optimization hyper-parameters to adjust to the few-shot learning scenario that is different from previously fixed hyper-parameters.  4) Another type of meta-learning in few-shot learning is to use external memories \cite{yoo2019coloring,santoro2016meta}.

All approaches mentioned above employ the spatial RGB images as the input of backbones. However, none of them make use of the frequency representations. In our work, we preprocess the RGB images with a DCT module to obtain the frequency representations and then input them to the backbone.

\subsection{Deep Learning in the Frequency Domain}

Frequency-domain based algorithms are widely used in network compression \cite{wang2018packing,chen2016compressing}, focusing on modifying the network to yield better efficiency. However, our DCT-based method focuses on increasing the classifying accuracy by reducing the input size with little modification of the model itself. The band-limited algorithm presented in \cite{dziedzic2019band} shows that the model focuses on leveraging lower-frequency components. However, this FFT-based method is more effective on larger kernels instead of the most commonly used smaller filters like $3\times 3$ and $1\times 1$ in most neural networks. \cite{gueguen2018faster} uses DCT coefficients during JPEG encoding which helps to improve the efficiency.  \cite{XuKai2020} shows that high-frequency channels of DCT could be removed without accuracy loss or even help to increase the classification accuracy in large-scale image classification. However, none of the above mentioned frequency-based algorithms try to jointly exploit the features from both the spatial and frequency domains. \cite{xu2019shifted} proposes to add spatial-spectral convolution blocks in convolution layers to learn more powerful representations, while it requires much revision on the network and extra computations. Moreover, it employs DCT in a different way from ours. Based on frequency analysis, 
we propose to explore its effect on few-shot learning by integrating features from both the spatial and frequency domains. 

\section{Methodology}

\subsection{Description on Few-shot Learning}

Considering a problem of few-shot classification, let $C_{source}$ and $C_{target}$ denotes the source classes and target classes, respectively, and these two classes are disjoint.  In the source classes $C_{source}$, we have abundant labeled samples as the training data $D_{source}$, while only a few labeled data $D_{target}$ are accessible for each class in the target classes $C_{target}$.  For a $k$-way $n$-shot learning problem, where we have $n$ labeled support samples for $k$ novel classes in $C_{target}$, our task is to classify a query sample into one of these $k$ support classes.

\begin{figure*}[tp]
\centering
\includegraphics[width=18cm,height=6 cm]{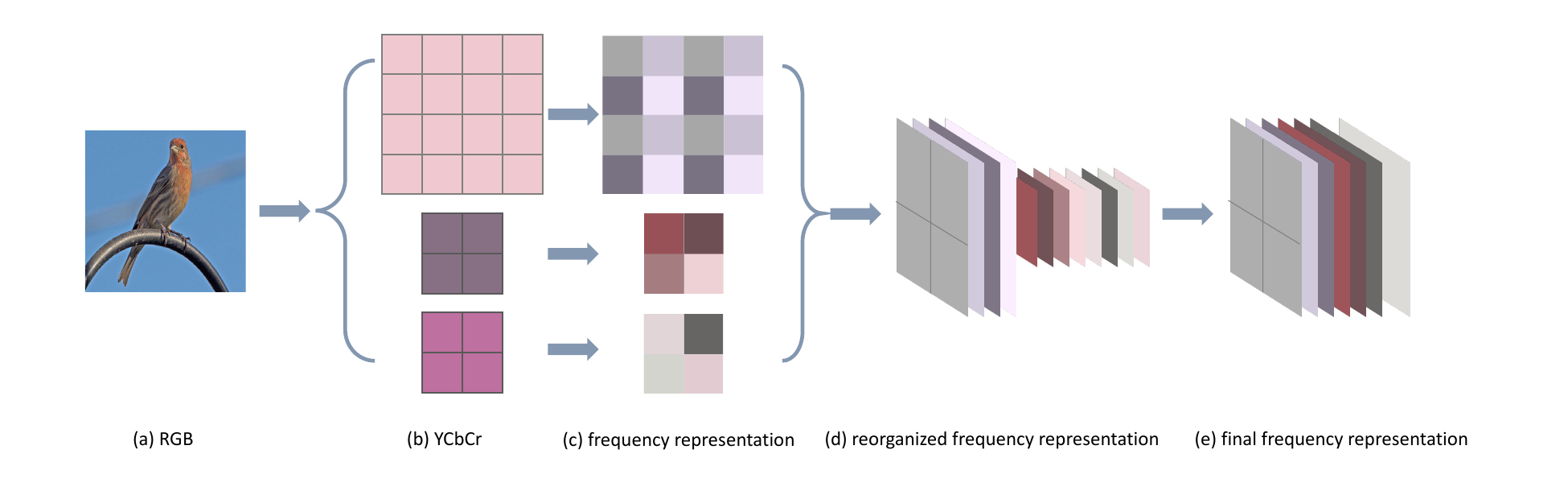}
\caption{{}{(a) images after preprocessing, e.g. $448\times448\times3$. (b) transform RGB to YCbCr images, Y: $448\times448$, Cr and Cb: $224\times224$. (c) frequency representation after DCT transformation with a filter, e.g. 8x8, (d) reorganize frequency representations by frequency channels, Y: $56\times56\times64$, Cb and Cr: $28\times28\times64$. (e) upsample all $Cr_{dct}$ and $Cb_{dct}$ frequency channels to the same size with $Y_{dct}$ channels and keep those frequency selected channels. Y: $56\times56\times16$, Cb and Cr: $28\times28\times4$.}}
\label{fig2}
\end{figure*}
\subsection{Discrete Cosine Transform (DCT)}

Inspired by the study \cite{XuKai2020}, we design the DCT pipeline as shown in figure \ref{fig2}. To generate the frequency representation, we perform some pre-processing first, including the standard transformation as illustrated in \cite{krizhevsky2017imagenet}, rotating, cropping, and translating, and obtain the image with size $S_{image}$ in step (a), e.g. $448\times 448$. Then, similar to the JPEG compression pipeline, we convert the high-resolution RGB images to YCbCr images, where we follow the 4:2:0 Chroma subsampling as shown in step (b) considering human vision system is more sensitive to brightness (Y) than color (Cr and Cb). After that, we divide each channel into $S_{dct}\times S_{dct}$ patches and perform $S_{dct}\times S_{dct}$ DCT transformation on each patch, where $S_{dct}$ is the size of the DCT filter, which is assigned at (e.g. $8\times 8$). In this way, we obtain an $8\times 8$ frequency representation for each $8\times 8$ patch in each YCbCr channel, with lower frequency in the left top corner and higher frequency in the right bottom corner as illustrated in step (c). Next, for each channel, we reshape these $8\times 8$ frequency patches to cubes by grouping the same frequency to one sub-channel and yield (d). Finally, from (d) to (e), we first select more impactful frequency sub-channels {}{obtained in} \cite{XuKai2020} from each YCbCr channel, concatenate them into one frequency cube and then upsample CrCb elements to the same size of Y element by interpolation, which would be the input of the following deep neural network. In (e), from left to right, it shows low to high-frequency elements for Y, Cb, and Cr respectively. 

The final input size after DCT module, the size of (f), would be $S_{image}/S_{dct}$, e.g. $448/8 = 56$, and the input size of ResNet is also $56\times 56$. So we could keep the network same size by adjusting $S_{image}$ and/or $S_{dct}$. In this way, we can handle images with a vast range of size to keep as much as information from the original images (e.g. when $S_{dct} = 16$ and the input size of the backbone is 56, we could process images with size $S_{image} = 56\times 16 = {}{896}$).  The DCT transform of an image can be denoted by:  %
\begin{equation}
D = TMT'
\end{equation}
\noindent
where $M$ is the $S_{dct}\times S_{dct}$ image patch after subtracting 128 for each pixel. And $T$ is the DCT transformation matrix determined by:
\begin{equation}
T_{i,j}=\left\{\begin{matrix}
\frac{1}{\sqrt{N}}, & i=0 \\ 
\sqrt{\frac{2}{N}}\cos\frac{(2j+1)i\pi}{2N}, & i>0
\end{matrix}\right.
\end{equation}

\subsection{Frequency Channels Selection}

\begin{figure}[t]
\begin{center}
    \includegraphics[width=3.2cm,height=3.2cm]{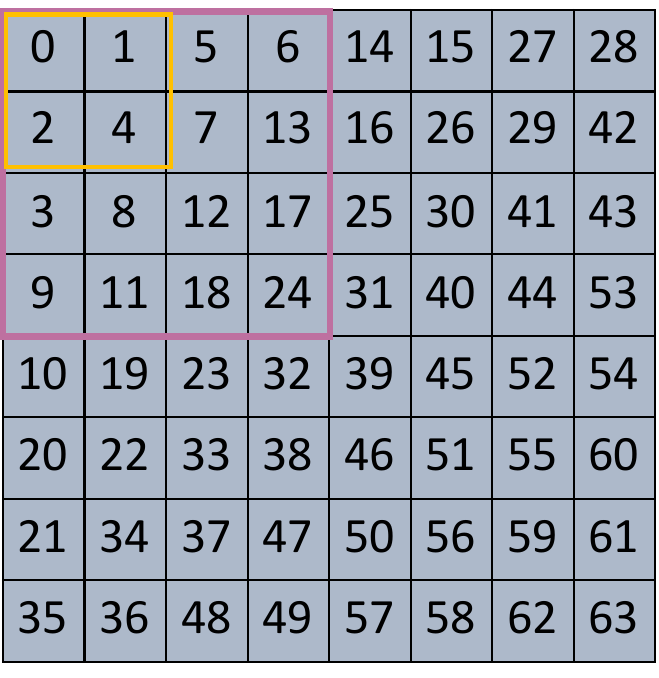}%
\end{center}
\caption{The $8 \times 8$ frequency index after DCT transformation. Each index represents one frequency component. Low frequencies lie in the top left corner and high frequencies are in the bottom right corner. We choose the top left square frequency elements as selected frequency channels. Specifically, $4\times 4$ square for Y channel and $2\times 2$ for the Cr and Cb channels. }
\label{fig3}
\end{figure}

According to the study \cite{XuKai2020}, low-frequency elements would be selected more frequently by designing a channel selection module carefully. In our implementation, we choose the following 24 channels {}{obtained in} \cite{XuKai2020} as shown in Figure \ref{fig3}: the top left $4\times 4$ square for Y channel and the top left $2\times 2$ for Cr and Cb channels.

\subsection{Network}

The framework of the proposed network is illustrated in Figure \ref{fig1}. The upper network, \enquote{CNN (s)} denotes a vanilla image classification network where we draw the input layers before the backbone from the whole CNN and\enquote{(s)} means the \enquote{spatial domain}. In the lower branch, \enquote{CNN (f)}, images will go through the DCT module first to obtain the frequency representation before being fed to the following CNN backbone, and \enquote{(f)} represents the \enquote{frequency domain}. Finally, we integrate the information from the two domains by concatenating the {}{normalized} features obtained from both networks to output the final classification score, which is denoted as the whole outer network, \enquote{CNN (s+f)}. 

\section{Experiments}

\subsection{Datasets and Setup} 

$\textbf{Datasets.}$  The proposed framework has been evaluated on three popular few-shot learning datasets: mini-ImageNet, CUB, and CIFAR-FS.  1) \noindent {\bf {mini-ImageNet}}. This is a popular few-shot learning dataset first proposed in  \cite{vinyals2016matching}, which samples 100 classes from the original ILSVRC-12 dataset \cite{russakovsky2015imagenet}.  For each class, it contains about 600 images. All images are $84\times 84$ RGB colored. In our experiment, we follow the split in  \cite{ravi2016optimization}, with 64 classes for training set, 16 classes for validation set, and 20 classes for testing set. Among these, classes from training, validation, and testing set are disjoint.  2) \noindent $\textbf{CUB}$. This dataset was introduced in  \cite{welinder2010caltech} and it contains 6,033 bird images, 130, 20, and 50 classes for training, validation, and testing, respectively. 3) \noindent $\textbf{CIFAR-FS}$. This dataset \cite{bertinetto2018meta} is obtained by randomly splitting 100 classes in CIFAR-100 \cite{krizhevsky2009learning} into 64 training classes, 16 validation classes, and 20 novel classes. All images in this dataset are of the size 32$\times$32.

$\textbf{Implementation details.}$  For CNN (s) where we input the backbones with images, the input samples are resized to $224\times 224$ for ResNet \cite{he2016deep} backbones and $84 \times 84$ for WRN-28-10 \cite{zagoruyko2016wide} on mini-ImageNet and CUB, $32 \times 32$ for CIFAR-FS.  For CNN (f), the frequency version, images are rescaled to $56\times S_{dct}$ and the input dimension varies from the number of selected frequency channels. For example, when we use 8$\times$8 DCT filters, the original images will be 448$\times$448. {}{To train (s+f), we train (s) and (f) separately first, and then fine-tune the classifier on the integrated features to get the final classifier for (s+f). This framework can also be trained end-to-end.}

\subsection{Application to Existing Learning Models}
In this section, we explore the generalization ability of the proposed approach to other few-shot learning frameworks.


\begin{table}[t]
\centering
\caption{Improvement after integrating features from both the spatial and frequency domains to existing methods on $\textit{mini}$ImageNet. The highest accuracy (\%) with $95\%$ confidence interval is highlighted. ${\#}$ and ${*}$  denotes results reported in  \cite{chen2019closer} and our reproduced results to the published ones respectively.}
\begin{tabular}{c|l|rr}
\toprule
\multirow{2}{*}{backbone} & \multirow{2}{*}{method} & \multicolumn{2}{c}{accuracy on $\textit{mini}$ImageNet} \\
& & 1-shot& 5-shot\\ \hline
\multirow{4}{*}{ResNet10}& MN \cite{vinyals2016matching}$^{\#}$ & 54.49$\pm$0.81 & 68.82$\pm$0.65 \\
 & MN (s)$^{*}$ &  52.98$\pm$0.21 & 72.41$\pm$0.16 \\
 & MN (f) &  55.98$\pm$0.20 & 74.17$\pm$0.16 \\ 
& MN (s+f) & \bf{57.32$\pm$0.21} & \bf{76.27$\pm$0.16} \\
&  & $\textit{+4.34}$&  $\textit{+3.86}$ \\ \hline

\multirow{4}{*}{WRN-28-10}  & S2M2$_{R} $\cite{mangla2020charting} &64.93$\pm$0.18 & 83.18$\pm$0.11 \\
& S2M2$_{R}$ (s)$^{*}$ &{}{63.09$\pm$0.17}  &{}{80.88$\pm$0.11} \\
& S2M2$_{R}$ (f)  & 63.03$\pm$0.18  &  80.80$\pm$0.11\\
& S2M2$_{R}$(s+f)  & {}{\bf{66.88$\pm$0.18}}  & {}{\bf{84.26$\pm$0.10}}  \\
&  & {}{$\textit{+3.79}$}&  {}{$\textit{+3.38}$} \\ \hline
\multirow{4}{*}{WRN-28-10} & PT+MAP \cite{hu2020leveraging}  & 82.92$\pm$0.26 & 88.82$\pm$0.13 \\
& PT+MAP (s)$^{*}$ &{}{80.73$\pm$0.24}&{}{87.81$\pm$0.13}  \\
& PT+MAP (f) &82.04$\pm$0.23 & 88.68$\pm$0.12  \\
& PT+MAP (s+f) &{}{\bf{84.81$\pm$0.22}} & {}{\bf{90.62$\pm$0.11}} \\
&  & {}{$\textit{+4.08}$}&  {}{$\textit{+2.81}$} \\ 
\bottomrule
\end{tabular}
\label{table5}
\end{table}

\begin{table*}[t]
\centering
\caption{Comparison with the state-of-the-art on mini-ImageNet, CUB, and CIFAR-FS. The highest accuracy (\%) is highlighted. * means results for miniImageNet and CUB-200-2011 datasets are from \cite{chen2019closer}, and results for CIFAR-FS are from \cite{bertinetto2018meta}.}
\begin{tabular}{l|l|cc||cc||cc}
\toprule
\multirow{2}{*}{method} & \multirow{2}{*}{backbone} & \multicolumn{2}{c||}{\bf{$\textit{mini}$ImageNet}} & \multicolumn{2}{c||}{\bf{CUB-200-2011}} & \multicolumn{2}{c}{\bf{CIFAR$-$FS}}  \\
& & 1-shot& 5-shot& 1-shot& 5-shot& 1-shot& 5-shot\\ \hline

ProtoNet$^{*}$ \cite{snell2017prototypical}& ConvNet&  50.37$\pm$0.83 & 67.33$\pm$0.67 & 66.36$\pm$1.00  & 82.03$\pm$0.59 & 55.5$\pm$0.70 & 72.0$\pm$0.60  \\
MAML$^{*}$ \cite{finn2017model} & ConvNet&  50.96$\pm$0.92 & 66.09$\pm$0.71 & 66.26$\pm$1.05 & 78.82$\pm$0.70 & 58.9$\pm$1.9 & 71.5$\pm$1.0\\
RelationNet$^{*}$ \cite{sung2018learning}& ConvNet&  51.84$\pm$0.88 & 64.55$\pm$0.70 & 64.38$\pm$0.94 & 80.16$\pm$0.64 & 55.0$\pm$1.0 & 72.0$\pm$0.60\\ \hline

S2M2$_{R}$ \cite{mangla2020charting}&WRN-28-10 & 64.93$\pm$0.18 &83.18$\pm$0.11 & 80.68$\pm$0.81 &90.85$\pm$0.44 & 74.81$\pm$0.19 &87.47$\pm$0.13 \\
AFHN \cite{li2020adversarial}& ResNet18  & 62.38$\pm$0.72 & 78.16$\pm$0.56 & 70.53$\pm$1.01 & 83.95$\pm$0.63 & - & -\\
DPGN \cite{yang2020dpgn}& ResNet12  & 67.77$\pm$0.32 & 84.60$\pm$0.43 & 75.71$\pm$0.47 & 91.48$\pm$0.33 & 77.90$\pm$0.50 & 90.20$\pm$0.40 \\
DeepEMD-sampling \cite{zhang2020deepemd}& ResNet12  & 68.77$\pm$0.29 & 84.13$\pm$0.53 & 79.27$\pm$0.29 & 89.80$\pm$0.51 & - & - \\
PT+MAP \cite{hu2020leveraging} & WRN-28-10 &82.92$\pm$0.26 & 88.82$\pm$0.13 &91.55$\pm$0.19 & 93.99$\pm$0.10 &87.69$\pm$0.23 & 90.68$\pm$0.15\\\hline
PT+MAP (s+f) (ours) &WRN-28-10  &{}{\bf{84.81$\pm$0.22}} & {}{\bf{90.62$\pm$0.11}} &{}{\bf{95.48$\pm$0.13}} & \bf{96.70$\pm$0.07} &{}{\bf{89.50$\pm$0.21}} & {}{\bf{92.16$\pm$0.15}} \\
\bottomrule

\end{tabular}
\label{table6}
\end{table*}

\subsubsection{Influence of the integrated features}

We integrate the proposed strategy to the following few-shot learning frameworks on mini-ImageNet: the Matching Network (MN) \cite{vinyals2016matching} (metric-based algorithm), S2M2$_{R}$ \cite{mangla2020charting} (pre-train and finetune), and PT+MAP \cite{hu2020leveraging} (post-process S2M2$_{R}$ features). The input is 84$\times$84 for the spatial branch to yield a fair comparison with previous results and 448$\times 448$ for the frequency domain. All images are pre-processed with data augmentation during training. The results are shown in Table \ref{table5}. It is evident that the proposed scheme (s+f) promotes the accuracy by about 2.8-4.3$\%$ in all cases compared with the original version (s) of the models. This shows that the integrating features from both domains work for different frameworks and its improvement is not limited to fine-tuning based or metric-learning based methods. Another interesting observation is that when we train S2M2$_{R}$ (f), the rotation loss decreases to less than 0.1 rapidly within a couple of epochs. This might because the rotation trick can work for spatial input while failed when it comes to frequency input, which might also explain that S2M2$_{R}$ (s) works better than S2M2$_{R}$ (f). This set of experiments demonstrates that the integrated features can work for different few-shot classification frameworks.

\subsubsection{Comparison with the state-of-the-art}

We compare the proposed network (s+f) with the state-of-the-art on the benchmarks. PT+MAP \cite{hu2020leveraging} proposes to leverage the learned features to Gaussian-like distribution and add it to the network S2M2$_{R}$ \cite{mangla2020charting}. Since the proposed strategy is designed in a preprocessing way, making it possible to combine it with any networks. We implement our method to PT+MAP and name it by adding \enquote{(s+f)} to the models we use. The images are resized to $84\times 84$ and $448\times 448$ for the spatial and frequency input respectively. 8$\times$8 DCT filter with static frequency channel selection is employed. 

The results are shown in Table \ref{table6}. It is evident that our method can increase the accuracy of the state-of-the-art by a large margin for all datasets we tested, including mini-Imagenet, CUB datasets, and CIFAR-FS. In all three datasets, PT+MAP achieves the best performance in terms of accuracy. For mini-Imagenet, our approach increases the best accuracy by {}{1.89}\% and {}{1.8}\% for 5-way 1-shot and 5-way 5-shot, respectively. For the CUB dataset, the accuracy is increased by {}{3.93}\% and 2.71\% respectively for the two tasks. Please note that for the CIFAR-FS dataset, the image size is small, only has 32$\times$32. However, we still observe {}{1.81}$\%$ and {}{1.48}$\%$ increases for the two tasks, respectively. 

\subsection{Ablation Study}

\subsubsection{Few-shot learning with DCT}


In this section, we first validate the effectiveness of DCT module and integrated network by comparing baseline++ (s), baseline++ (f) and baseline++ (s+f) w/o data augmentation during training on $\textit{mini}$ImageNet, where baseline++ \cite{chen2019closer} uses all images from base classes to pre-train and then fine-tune with a few support samples from testing set before testing. The only difference between the baseline \cite{qi2018low} and baseline++ is that the baseline uses a linear classifier while baseline++ calculates cosine distance. The backbone we choose for feature extraction is ResNet 34. Both 5-way 1-shot and 5-way 5-shot image classification tasks are evaluated. {}{For data augmentation during training, we performed random crop, left-right flip and color jitter as in the paper \cite{chen2019closer}}

\begin{table}[pb]
\centering
\caption{Results on $\textit{mini}$ImageNet with different inputs as shown in Figure \ref{fig1}. The backbone is ResNet34. Top left square 24 channels {}{as illustrated in Figure \ref{fig3}} are selected for frequency versions. The highest accuracy (\%) is highlighted.}
\begin{tabular}{l|c|cc}
\toprule
\multirow{2}{*}{method} & {}{data} & \multicolumn{2}{c}{$\textit{mini}$ImageNet} \\
&{}{augmentation} &5-way 1-shot& 5-way 5-shot\\ \hline
baseline++ (s) & {}{No} & 48.54$\pm$0.17 & 61.58$\pm$0.13 \\
baseline++ (f) & {}{No} & 53.27$\pm$0.19 & 65.65$\pm$0.13 \\
baseline++ (s+f) & {}{No} & {}{\bf{54.76$\pm$0.18}} & {}{\bf{68.76$\pm$0.13}} \\\hline
baseline++ (s) & {}{Yes} & 57.94$\pm$0.18 & 73.98$\pm$0.13 \\
baseline++ (f) & {}{Yes} & 59.22$\pm$0.18 & 76.58$\pm$0.13 \\
baseline++ (s+f) & {}{Yes} & {}{\bf{62.75$\pm$0.18}} & {}{\bf{79.73$\pm$0.12}} \\
\bottomrule
\end{tabular}
\label{table1}
\end{table}

The experimental results are shown in Table \ref{table1}, from which we can see that the accuracy of baseline++ (f) is higher than baseline++ (s) by 4.73$\%$ and 1.28$\%$ respectively for 5-way 1-shot classification task without and with data augmentation during the first training phase, 4.07$\%$ and 2.6$\%$ for the 5-way 5-shot task. The baseline++ (s+f) in {}{all} cases further increases the classification accuracy by 1.5-3.5 $\%$, showing that baseline++ (f) is not just an improvement of  baseline++ (s) but a complementary method of it and learning from both the spatial and frequency domain could increase the classification accuracy.

\subsubsection{Influence of original information quantity}

For the DCT branch, the network has more flexibility to choose image size with the existence of the DCT module, even take larger images compared with the spatial version.  To explore the effect of information quantity the frequency branch takes, we conducted experiments on $\textit{mini}$ImageNet. The backbone of the frequency branch is ResNet 10 to save time and data augmentation is employed during training. The results are tabulated in Table \ref{table2}. For the first two parts in the table, we preprocess images for baseline++ (f) {}{with the same data augmentation method} as the baseline (s) to generate 84$\times$84 and 224$\times$224 images. Then, we upsample these images to 448$\times$448 to exploit the DCT module. For the bottom part in this table, we resize the images directly to 448$\times$448 to include more original information. 

From the table, we can see that, when baseline++ (s) and baseline++ (f) utilize the same information from the original images, baseline++ (s) performs better than baseline++ (f) when the image is $84\times 84$ but worse for image size $224\times 224$, which means (f) holds the potential to perform better with fewer parameters than (s) when inputting enough information. When we input baseline++ (f) with more information than baseline (s), 448 instead of 224, the accuracy for (f) gets slightly improved, 0.9 \% for 1 shot and 0.06 \% for 5 shot. We think the reason is the information quantity is more and more approaching the amount needed by the current frequency backbone. Moreover, in all cases baseline++ (s+f) performs better than both baseline++ (s) and baseline++ (f), which means that the spatial and frequency representation are complementary to each other. To conclude, we can use larger images (if we could access them) to increase the accuracy by using the DCT module, and integrated features can always improve the performance no matter whether the frequency branch could access larger images compared to the spatial branch. 


\begin{table}[tp]
\centering
\caption{Results on $\textit{mini}$ImageNet with different information quantity supplied to the frequency channel as shown in Figure \ref{fig1}. The backbone is ResNet 10 and data augmentation when training is implemented. The highest accuracy (\%) is highlighted.}
\begin{tabular}{l|c|cc}
\toprule
\multirow{2}{*}{method} & image & 
\multicolumn{2}{c}{$\textit{mini}$ImageNet} \\
& size &5-way 1-shot& 5-way 5-shot\\ \hline
baseline++ (s) & 84 & 52.32$\pm$0.17 & 68.24$\pm$0.14 \\
baseline++ (f) & 84$\rightarrow$448 & 49.47$\pm$0.17 & 65.64$\pm$0.12 \\
baseline++ (s+f) & N/A & {}{\bf{56.32$\pm$0.17}} & {}{\bf{75.70$\pm$0.13}} \\\hline
baseline++ (s) & 224 & 57.52$\pm$0.17 & 75.56$\pm$0.13  \\
baseline++ (f) & 224$\rightarrow$448 & {}{58.71$\pm$0.17} & {}{76.55$\pm$0.12} \\
baseline++ (s+f) & N/A & {}{\bf{62.23$\pm$0.18}} & {}{\bf{80.08$\pm$0.12}} \\\hline
baseline++ (s) & 224 & 57.52$\pm$0.17 & 75.56$\pm$0.13  \\
baseline++ (f) & 448 & 59.61$\pm$0.18 & 76.61$\pm$0.12  \\
baseline++ (s+f) & N/A & {}{\bf{62.30$\pm$0.18}} & {}{\bf{79.93$\pm$0.11}} \\

\bottomrule
\end{tabular}
\label{table2}
\end{table}

\subsubsection{Different DCT filters and selected channels}
In this experiment, we explore the effect of different sizes of DCT filters, 2, 4, 6, and 8, and different selected channels, 24 channels and all frequency channels as shown in Table \ref{table3}. The backbone is ResNet18. For (f) version, the number of channels is 24 and $S_{dct}\times S_{dct}\times 3$ when we select 24 and all channels respectively, e.g. when the DCT filter size is 4, $S_{dct} = 4$, the number of all channels will be $4\times 4\times 3 = 48$, where $4\times 4$ is the size of DTC module and 3 is from Y, Cr and Cb channels. For the spatial branch, we use all 224 as the image size. For the DCT branch, we resize images to $56\times S_{dct}$ directly, e.g. if $S_{dct} = 4$, the input is rescaled to $56\times 4 = 224$. Experiments with and without data augmentation during training are evaluated.

\begin{table*}[ht]
\centering
\caption{The effect of different sizes of DCT filters and whether we select frequency channels with backbone ResNet18 on $\textit{mini}$ImageNet.  The number of channels denotes the channels before the backbone with the input layers removed, e.g. for ResNet, the input before the basic backbone is $56\times 56\times 64$. The highest accuracy (\%) is highlighted.}
\begin{tabular}{l|c|c|c|cc}
\toprule
\multirow{2}{*}{method} &\multirow{2}{*}{train$_{aug}$} & {DCT} & \multirow{2}{*}{channels} & \multicolumn{2}{c}{accuracy on $\textit{mini}$ImageNet} \\
& & filter size & &5-way 1-shot& 5-way 5-shot\\ \hline
baseline++ (s) & False & - & 64 & 48.40$\pm$0.17 & 62.94$\pm$0.13 \\
baseline++ (f) & False & 2 & all (12)& 49.05$\pm$0.17 & 65.45$\pm$0.13 \\
baseline++ (f) & False & 4 & all (48)& 49.67$\pm$0.18 & 65.07$\pm$0.13 \\
baseline++ (f) & False & 6 & all (108)& 49.70$\pm$0.17 & 64.51$\pm$0.13 \\\hline
baseline++ (s) & True & - & 64 & 56.48$\pm$0.17 & 74.00$\pm$0.13 \\

baseline++ (f) & True  & 2 & all (12)& 57.79$\pm$0.17 & 75.50$\pm$0.12 \\
baseline++ (f) & True  & 4 & all (48)& 58.41$\pm$0.17 & 76.01$\pm$0.12 \\
baseline++ (f) & True  & 6 & all (108)& \bf{58.98$\pm$0.17} & 75.39$\pm$0.12 \\\hline\hline

baseline++ (f) & False & 4 & 24 & 50.11$\pm$0.17 & 63.91$\pm$0.13 \\
baseline++ (f) & False & 6 & 24& 50.72$\pm$0.18 & 64.86$\pm$0.13 \\
baseline++ (f) & False & 8 & 24& 51.04$\pm$0.18 & 65.76$\pm$0.13 \\\hline

baseline++ (f) & True  & 4 & 24 & 58.02$\pm$0.18 & 75.73$\pm$0.13 \\
baseline++ (f) & True  & 6 & 24& 57.74$\pm$0.17 &75.66$\pm$0.12 \\
baseline++ (f) & True  & 8 & 24& 58.25$\pm$0.18 & \bf{76.23$\pm$0.13} \\
\bottomrule
\end{tabular}
\label{table3}
\end{table*}

According to Table \ref{table3}, all baseline++ (f) methods outperform their corresponding baseline++ (s) version without the DCT module, even when the DCT filter size is as small as 2$\times$2. When all frequency channels are employed, we find the accuracy is increased with the increase of the size of the DCT filter no matter there is a data augmentation or not for the 5-way 1-shot classification. However, the increase is very small, and for 5-way 5-shot task, we do not observe this trend, and the baseline++ (f) with $S_{dct} = 2$ even outperforms other filter sizes when there is no data augmentation. On the other hand, when only the top left 24 frequency channels are employed, the accuracy increases mildly with that of the filter sizes for both 5-way 1-shot and 5-way 5-shot.

From these observations, we can see that the filter size has little influence on the few-shot classification when we perform static frequency channel selection. The influence is neglectable in comparison with the influence of data augmentation. In this experiment, DCT filter size 6 with all channels and data augmentation achieves the best performance for the 5-way 1-shot task, and DCT filter size 8 with 24 channels and data augmentation achieves the best performance for the 5-way 5-shot task. However, the increase is not significant. In practice, we can simply choose a small filter size to save the computation cost.

\begin{table}[t]
\centering
\caption{Results on $\textit{mini}$ImageNet with different backbone networks when we implement different input versions, (s), (f) and (s+f), as shown in Figure \ref{fig1} to baseline++. The highest accuracy (\%) is highlighted.}
\begin{tabular}{c|l|rr}
\toprule
\multirow{2}{*}{backbone} & \multirow{2}{*}{method} & \multicolumn{2}{c}{accuracy on $\textit{mini}$ImageNet} \\
& &1-shot&  5-shot\\ \hline
\multirow{3}{*}{ResNet10} 
& baseline++ (s) &  57.52$\pm$0.17 & 75.56$\pm$0.13 \\
& baseline++ (f) &  59.61$\pm$0.18 & 76.61$\pm$0.12 \\ 
& baseline++ (s+f) & {}{\bf{62.30$\pm$0.18}} & {}{\bf{79.93$\pm$0.12}}\\
&  & {}{$\textit{+4.78}$}&  {}{$\textit{+4.37}$} \\ \hline
\multirow{3}{*}{ResNet18}  & baseline++ (s) & 56.48$\pm$0.17 & 74.00$\pm$0.13 \\
& baseline++ (f)  & 58.52$\pm$0.18 & 76.23$\pm$0.13 \\
& baseline++ (s+f)  & {}{\bf{61.66$\pm$0.18}} & {}{\bf{79.70$\pm$0.12}} \\
&  & {}{$\textit{+5.18}$}&  {}{$\textit{+5.70}$} \\ \hline
\multirow{3}{*}{ResNet34} & baseline++ (s) & 57.94$\pm$0.18 & 73.98$\pm$0.13 \\
& baseline++ (f)  & 59.22$\pm$0.18 & 76.58$\pm$0.13 \\
& baseline++ (s+f) & {}{\bf{62.75$\pm$0.18}} & {}{\bf{79.73$\pm$0.12}} \\
&  & {}{$\textit{+4.81}$}&  {}{$\textit{+5.75}$} \\
\bottomrule
\end{tabular}
\label{table4}
\end{table}


\subsubsection{Integrated features with different backbone}

In this section, to verify the impact of integrated features under different backbones, we implement different versions of baseline++, (f) and (s+f), with the backbone ResNet10, ResNet18, and ResNet34 as shown in Table \ref{table4}. Image size for (s) and (f) is 224 and 448 respectively. $8\times 8$ DCT filters and static channel selection are implemented. In all cases, baseline++(f), when we preprocess images with the DCT module, outperforms all baseline++ (s) when there is no DCT module. Furthermore, in both tasks (5-way 1-shot and 5-way 5-shot classification) and for all backbones (ResNet10, ResNet18, and ResNet34), the baseline++ (s+f) version achieves the best performance compared with baseline++(f) and baseline++(s), and the accuracy is improved by a margin of 4-6$\%$ compared with their vanilla versions, baseline++(s).  This further verifies that learning from the frequency domain is a complementary method of learning from the spatial domain and integrating both features could further increase the classification accuracy. The experiment also demonstrates the effectiveness of the proposed approach on different backbone networks.


\begin{figure}[tb]
\begin{center}
\subfloat[]{
    \includegraphics[width=3.5 cm,height=3.5 cm]{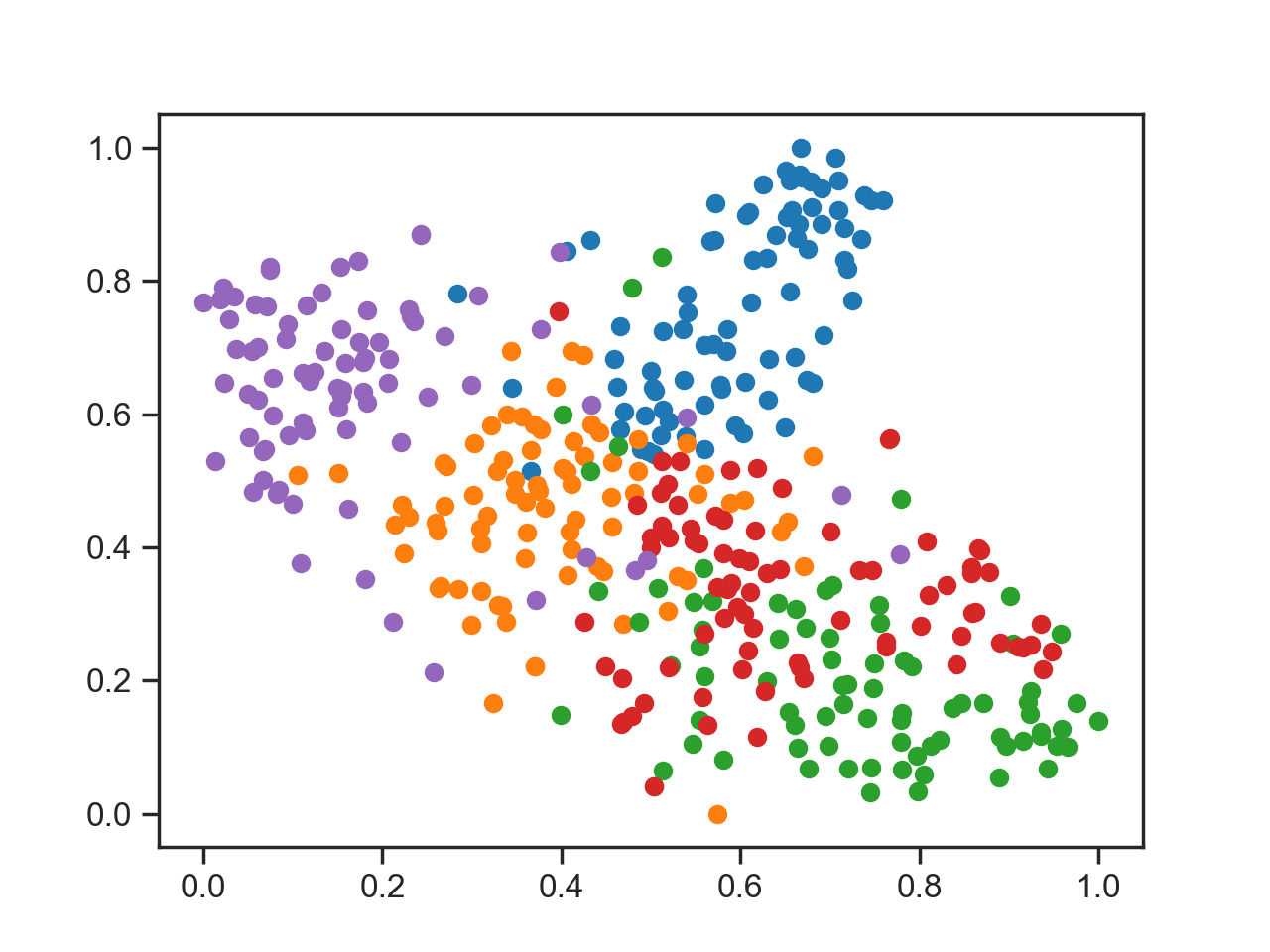} }%
\qquad
\subfloat[]{
    \includegraphics[width=3.5 cm,height=3.5 cm]{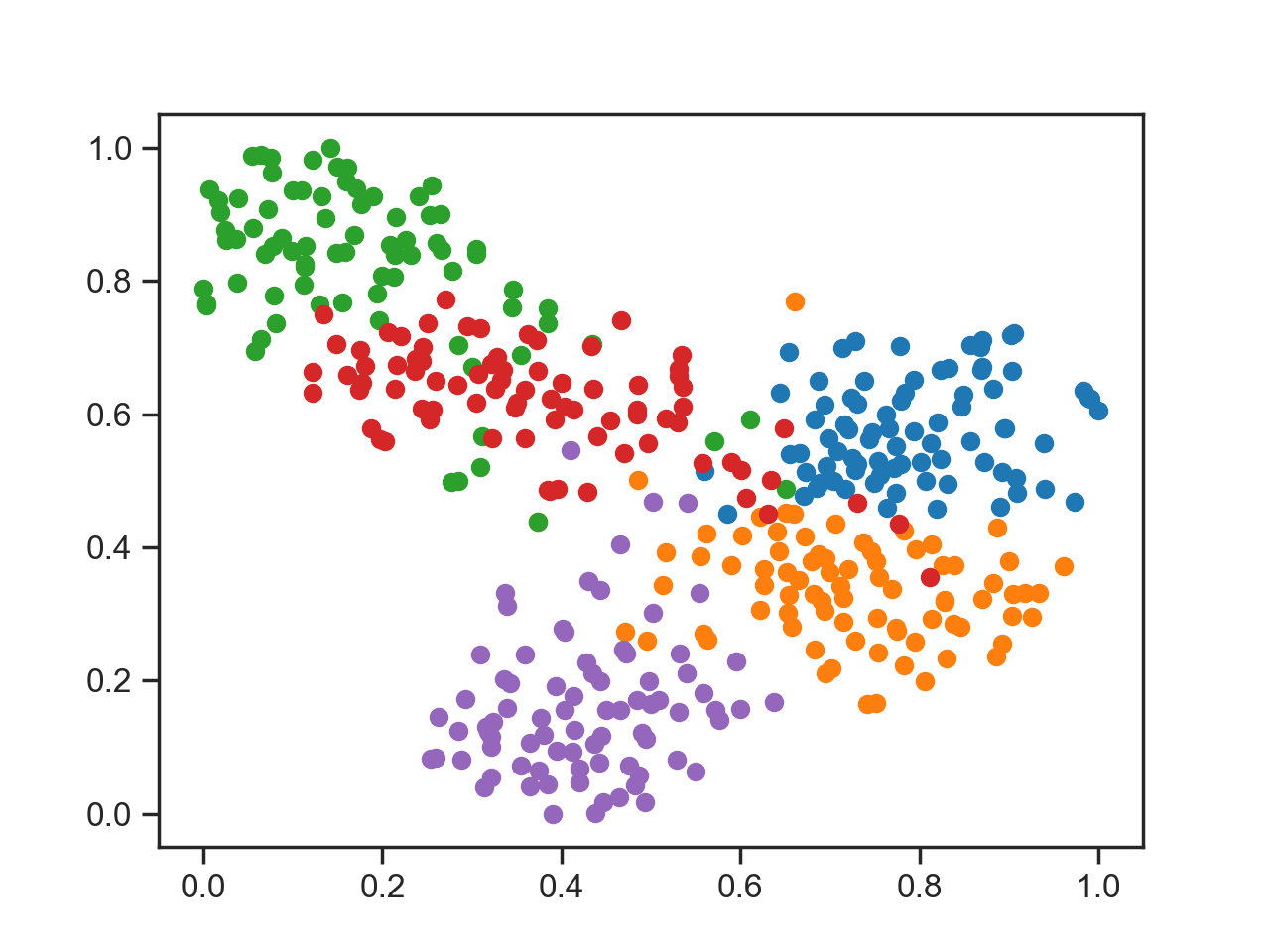}}%
\end{center}
\caption{t-SNE feature visualization results. (a): baseline++(s).  (b): baseline++(s+f).}
\label{fig4}
\end{figure}

\subsection{Feature Visualization}

To visually understand the feature we learned from the whole framework, the t-SNE visualization \cite{maaten2008visualizing} is shown in Figure \ref{fig4}. The features for (a) and (b) are learned from baseline++ with backbone ResNet 10, specifically baseline++(s) and baseline++(s+f). 5 novel categories are selected randomly and 80 samples for each class are employed. It can be observed from Figure \ref{fig4} that the clustering results by integrating features from both the spatial and frequency domains are more compact than those from only the spatial domain, which further verify the effectiveness of integrating both domains in improving the clustering ability. 

\section{Conclusion}

In this paper, we have proposed to apply the DCT pre-processing technique to generate the frequency information of images and integrate the representations from both the spatial and frequency domains to increase the performance of few-shot classification. Through extensive experiments, we have demonstrated that the frequency information is complementary to feature representation, and integrating the features learned from both the spatial and frequency domains can significantly increase the performance of few-shot learning. The proposed strategy can act as a plug-in module for other few-shot learning models to increase their classification accuracy.


\section*{Acknowledgement}

The work was supported in part by The National Aeronautics and Space Administration (NASA) under grant no. 80NSSC20M0160.

%

\balance
\end{document}